\documentclass{llncs}

\usepackage{makeidx} 
\usepackage[misc,geometry]{ifsym} 
\usepackage{amsmath} 
\usepackage{bm} 
\usepackage{graphicx} 
\graphicspath{ {.} }
\usepackage{array} 
\usepackage[table,xcdraw]{xcolor} 
\usepackage{color} 
\usepackage{afterpage} 
\usepackage[labelfont=bf]{caption} 
\usepackage{footmisc} 
\usepackage{fixltx2e} 
\usepackage{cite} 
\usepackage[colorlinks]{hyperref} 

\usepackage{etoolbox}
\makeatletter
\let\llncs@addcontentsline\addcontentsline
\patchcmd{\maketitle}{\addcontentsline}{\llncs@addcontentsline}{}{}
\patchcmd{\maketitle}{\addcontentsline}{\llncs@addcontentsline}{}{}
\patchcmd{\maketitle}{\addcontentsline}{\llncs@addcontentsline}{}{}
\makeatother
\usepackage{hyperref}
\usepackage{bookmark}

\begin{document}

\title{SqueezeJet: High-level Synthesis Accelerator Design for Deep Convolutional Neural Networks}

\titlerunning{SqueezeJet: High-level Synthesis Accelerator Design for Deep Convolutional Neural Networks}  

\author{Panagiotis~G.~Mousouliotis\textsuperscript{(\Letter)},
Loukas~P.~Petrou}

\authorrunning{Panagiotis G. Mousouliotis et al.} 

\tocauthor{Panagiotis G. Mousouliotis, Loukas P. Petrou}

\institute{Division of Electronics and Computer Engineering,\\
Department of Electrical and Computer Engineering, Faculty of Engineering,\\
Aristotle University of Thessaloniki, 54124 Thessaloniki, Greece\\
\email{pmousoul@ece.auth.gr}\\
\email{loukas@eng.auth.gr}}

\maketitle

\begin{abstract}

Deep convolutional neural networks have dominated the pattern recognition scene by providing much more accurate solutions in computer vision problems such as object recognition and object detection. Most of these solutions come at a huge computational cost, requiring billions of multiply-accumulate operations and, thus, making their use quite challenging in real-time applications that run on embedded mobile (resource-power constrained) hardware. This work presents the architecture, the high-level synthesis design, and the implementation of SqueezeJet, an FPGA accelerator for the inference phase of the SqueezeNet DCNN architecture, which is designed specifically for use in embedded systems. Results show that SqueezeJet can achieve 15.16 times speed-up compared to the software implementation of SqueezeNet running on an embedded mobile processor with less than 1\% drop in top-5 accuracy.


\end{abstract}

\section{Introduction}
\label{sec:intro}

Since the impressive results of AlexNet deep convolutional neural network (DCNN) in the Image-Net Large-Scale Vision Recognition Challenge (ILSVRC) in 2012 \cite{kriz}, DCNN research activity has seen exponential growth with the trend being deeper architectures accompanied by higher accuracies \cite{szeg,he}. Following this trend, research in DCNN FPGA accelerators provides solutions that use high-end costly FPGA devices and aim at the datacenter rather than the mobile applications \cite{zhang,mota,ovtc}. An exception to the-most-accurate-network trend in the DCNN architecture research, is SqueezeNet\footnote{In this work, SqueezeNet refers to SqueezeNet v1.1} (SqN) \cite{iand1,iand}, an AlexNet-level accuracy architecture which reduces dramatically the number of MACs and network parameters, requiring half of the MACs and fifty times less parameters compared to AlexNet. Even though the SqN DCNN architecture is more suitable than others for use in embedded mobile applications, it is still computationally very demanding and cannot be used in applications running on an embedded mobile processor.
\par
The contribution of this work is the design of SqueezeJet (SqJ), a small FPGA convolutional (conv) layer accelerator for SqN, that can be used as a coprocessor to an embedded mobile processor and enable the development of mobile computer vision (CV) applications. Specifically, the SqJ design: (1) deals with the challenge of the implementation of a single accelerator for multiple conv layers with variable input arguments, (2) implements streaming input/output (I/O) interfaces which, after the initialization phase, consume and produce data pixel-by-pixel\footnote{A pixel is comprised by all the channels at a specific $(x, \, y)$ location in the future map volume (see Figure \ref{fig:conv3d}).}, (3) uses a sophisticated hardware (HW) mechanism, which mimics software (SW) pointers to the rows of a two-dimensional array, taking advantage of the spatial locality of data and minimizing unnecessary data movement, (4) presents the possibilities of high-level synthesis (HLS) design by using the Xilinx Vivado HLS (VHLS) tool, (5) is implemented on a low-end FPGA system on chip (SoC) device, the Xilinx XC7Z020, using the Xilinx SDSoC tool, and (6) it achieves 80.29\% ILSVRC12 top-5 accuracy when it is used for the inference phase of SqN. To the best of the authors' knowledge, the current work presents the first low-end FPGA SoC (XC7Z020) DCNN implementation which achieves 80.29\% ILSVRC12 top-5 accuracy.

The rest of this paper is organized as follows: Section \ref{sec:sota} presents related work. Section \ref{sec:conv} is an introduction to the conv layer's operation. Section \ref{sec:sqj} presents the architecture, the HLS design, and the implementation of the SqJ accelerator. Section \ref{sec:perf} shows results related to the performance, the accuracy, and the power consumption of SqJ. Finally, Section \ref{sec:con} concludes the paper and proposes future work.

\section{Related work}
\label{sec:sota}

Works related to DCNN FPGA accelerators can be classified into two main categories; those which accelerate only the conv layer and those which accelerate two or more layer types of a DCNN.
\par
\textbf{Conv layer accelerators:}  \textit{Zhang et al.} \cite{zhang} designed an architecture template for the conv layer using loop tiling, loop arrangement based on data dependencies, computation optimizations (loop unrolling and pipelining), and optimizations for efficient data reuse. Using the parameters of the template and the roofline model, they performed design space exploration (DSE) and found the optimal solution which defined the parameters of their accelerator. A similar approach is followed by \textit{Motamedi et al.} \cite{mota} starting with a completely different architectural template. Specifically, they designed their template to take advantage of all the possible forms of parallelism; intra/inter-kernel and inter-output. They eventually used the design parameters and proceeded as in the aforementioned work. Both of these works use DSE to minimize the execution time of the accelerator and 32-bit floating-point arithmetic. 
\par
\textbf{Multi-layer accelerators:}  \textit{Qiu et al.} \cite{qiu} developed a dynamic-precision data quantization flow and designed a dynamic-precision 16-bit fixed-point accelerator which is capable of accelerating conv, fully connected (FC), and pooling layers. Their implementation is used to accelerate the VGG16-SVD DCNN, which is the VGG16 DCNN with reduced weight matrices for the FC layers; SVD is used for the weight matrix reduction. This accelerator also uses a huge amount of FPGA resources to accelerate one of the most computational demanding DCNNs, requiring 15470 million MACs for a single forward pass. \textit{Gschwend} \cite{gschw} converted all the layers, except the last global pooling layer, of the SqueezeNet v1.0 DCNN architecture to conv layers and accelerated, using floating-point arithmetic, the new DCNN, called ZynqNet, using VHLS.  \textit{Gokhale et al.} \cite{gok} designed and implemented nn-X, a complete low-power system for DCNN acceleration composed from a host processor, a coprocessor, and external memory. The coprocessor consists of an array of processing elements which can perform convolution, sub-sampling, and non-linear functions.  \textit{Ma et al.} \cite{ma} designed an accelerator that supports conv, pooling and fully-connected layers by following a strategy that minimizes computing latency, partial sum storage, access of on-chip buffer, access of external memory, and uses loop optimization techniques. Their accelerator uses 8-16 bit dynamic fixed point arithmetic and it is evaluated by accelerating the VGG-16 DCNN. 
\par
SqJ is a conv layer accelerator and it uses fixed-point arithmetic for both parameters (8 bits) and activations (16 bits), which results in considerable savings in both the resources and the power consumption compared to floating-point implementations \cite{zhang,mota,gschw}. Furthermore, even though works in \cite{qiu,gok,ma} use fixed-point arithmetic, they require large costly FPGA devices for their implementation.

\section{Convolutional layer basics}
\label{sec:conv}

The conv layer of a DCNN can be described by:

\begin{equation}
\label{eq:conv}
\begin{gathered}
FM_{o}(y_o ,\, x_o ,\, c_o)=\\ {\sum_{k_h=0}^{K_h-1}\sum_{k_w=0}^{K_w-1}\sum_{c_i=0}^{C_i-1}
FM_i((y_o \cdot S+k_h) ,\, (x_o \cdot S+k_w) ,\, c_i) \cdot W(c_o ,\, k_h ,\, k_w ,\, c_i)}\\
+B(c_o),
\end{gathered}
\end{equation}
where $FM_o$, $FM_i$ are the output and the input future maps (fmaps) respectively, and $W$, $B$ are the weight and bias parameters respectively. The $y$, $x$, $c$, represent the vertical, the horizontal, and the channel dimensions of the fmaps, $S$ is the stride, and $k_h$, $k_w$ are the vertical and horizontal dimensions of the kernel\footnote{In this work, kernel has the same meaning as filter.}.

\begin{figure}[]
    \centering
    \fbox{ \includegraphics[scale=0.22]{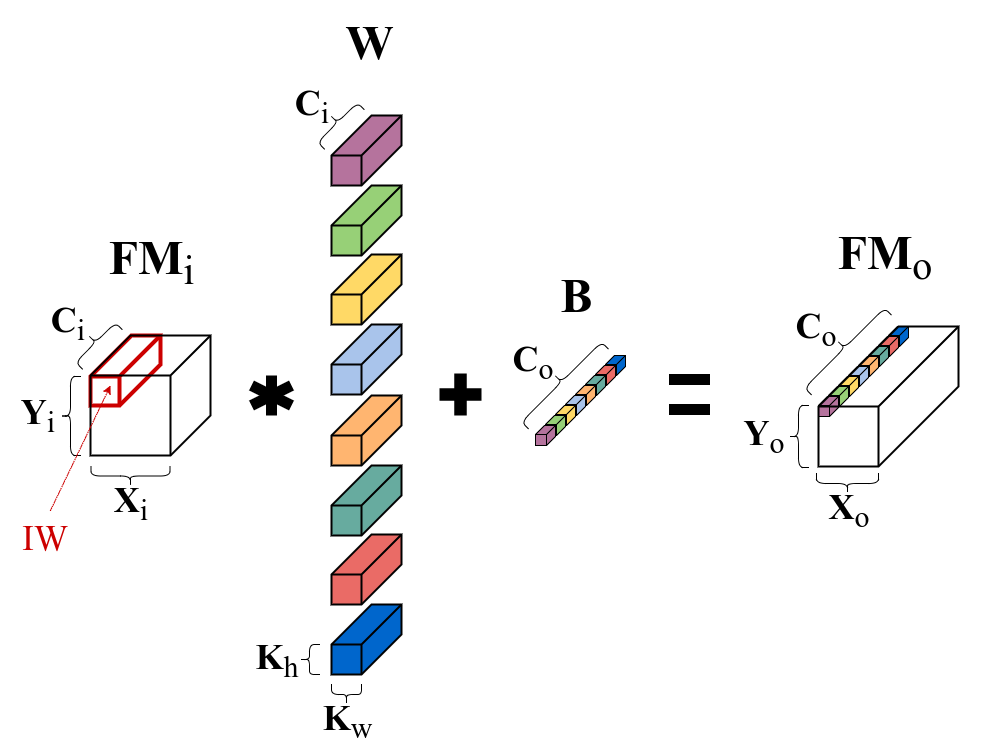} }
    \caption[]{Calculation of the channels of the first pixel of $FM_o$. The number of 3D kernels is equal to $C_o$, the number of output channels.}
    \label{fig:conv3d}
\end{figure}

\par
The second line in Equation \ref{eq:conv} represents a 3D convolution between $FM_i$, and $C_o$ number of 3D kernels, the weight parameters. To calculate the first output channel of the first output pixel of $FM_o$, an input window $IW$ of the $FM_i$, of size $IW[K_h][K_w][C_i]$ is multiplied element-wise with kernel $W[0][K_h][K_w][C_i]$ and all partial results are accumulated to a single value. This value is then added to the respective bias term, $B(0)$, to produce the first output channel of the first output pixel of $FM_o$. This procedure is depicted in Figure \ref{fig:conv3d}, which shows the calculation of all the channels of the the first pixel of $FM_o$. To calculate all the elements of $FM_o$, the $IW$ is moved vertically and horizontally by $y_o \cdot S$ and $x_o \cdot S$ respectively, and the above procedure is repeated. The resulted size of each $Y$, $X$ dimension of the $FM_o$ is calculated by:

\begin{equation}
\label{eq:fmoxy}
\begin{gathered}
Y_o = (Y_i - K_h + 2 \cdot P)/S + 1 \\
X_o = (X_i - K_w + 2 \cdot P)/S + 1,
\end{gathered}
\end{equation}
where $P$ denotes the number of pixels added for padding the $FM_i$. In all the practical cases $Y_i=X_i$ and $K_h=K_w$.

\par
An activation function always follows a conv layer. Thus, it is convenient, from an implementation point of view, to include the activation layer in the conv layer. In this case, the output of the conv layer becomes:

\begin{equation}
\label{eq:fmoa}
FM_{o,a}(y_o ,\, x_o ,\, c_o)=f(FM_{o}(y_o ,\, x_o ,\, c_o)),
\end{equation}
where $f()$ is the activation function used in the specific DCNN, e.g. the Rectified linear unit (ReLU) described by:

\begin{equation}
\label{eq:relu}
f(x)=max(0, \, x)
\end{equation}

\par
The accelerator described in the next section, accelerates this fused convolution-activation layer with output given by Equation \ref{eq:fmoa}.

\section{The SqueezeJet accelerator}
\label{sec:sqj}

SqN is a DCNN architecture focused in reducing the network parameter count for a given accuracy. Specifically, SqN achieves AlexNet-level accuracy with fifty times less parameters, making its model sufficiently small to be stored in on-chip FPGA memories and removing the need for off-chip memory access. For an FPGA accelerator, such as SqJ, implemented on a device with a few Mbits of block RAM (BRAM) resources, this means that the parameters (weight and bias values) of a single layer can fit in the BRAMs. Thus, for the calculation of the $FM_o$ of a specific conv layer by an accelerator, the following procedure is required: the parameters are brought from off-chip memory and stored to BRAMs, the $FM_i$ is streamed from off-chip memory in the accelerator, the calculation of $FM_o$ pixel(s) takes place, and the resulting $FM_o$ pixel(s) are streamed back to the off-chip memory. Having the layer's parameters stored on-chip is a big advantage as they will be reused for the calculation of each pixel of $FM_o$. 

Following the architecture principle ``make the common case fast'', SqJ is designed to accelerate conv layers described by Equation \ref{eq:fmoa} with stride limited to one; it can be used for the acceleration of all the SqN conv layers except the first one, which can be implemented as a distinct module. All SqN conv layers, except the first one, share the following common characteristics: (1) a stride equal to 1, (2) an input channel dimension with a greatest common divisor (GCD) equal to 16 and (3) an output channel dimension which is divisible by a power of 2. SqJ uses all these three characteristics to accelerate a conv layer; the first SqN conv layer does not have characteristics (1) and (2). Implementing SqJ to support the first SqN conv layer would significantly degrade the acceleration of the other 17 conv layers (25 conv modules) of SqN.

This section describes the the architecture, the high-level synthesis design, and the implementation of SqJ.

\subsection{Architecture}
\label{sec:sqj_arch}

\begin{figure}[]
    \centering
    \fbox{ \includegraphics[scale=0.25]{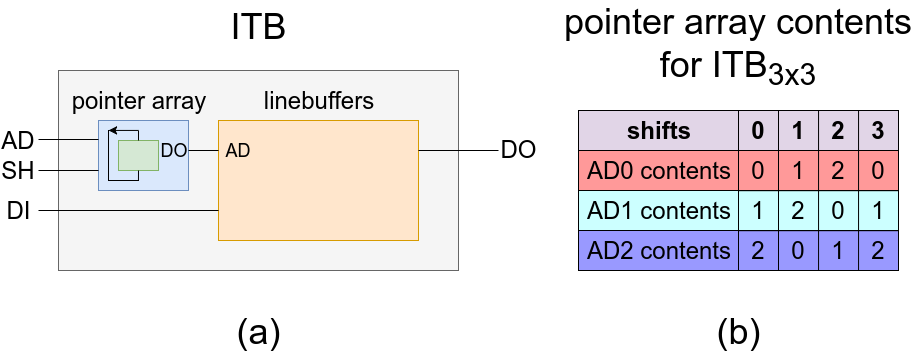} }
    \caption[]{$ITB$: \textbf{(a)} schematic and \textbf{(b)} pointer array content of $ITB_{3 \times 3}$ after a number of shifts. AD denotes memory address, SH denotes a shift signal, and DI, DO denote data input and output respectively.}
    \label{fig:itb}
\end{figure}

\textbf{Data organization:} The data organization of all the convolution array arguments is shown in Equation \ref{eq:conv}. This data organization is imposed by the 3D convolution operation; it is necessary to read all the input channels of the $IW$ pixels in order to be able to calculate a single output channel. Because SqJ accelerates 3D convolutions, the design of a streaming architecture is not possible, but it is possible to design the accelerator to use streaming I/O interfaces.
\\
\\
\textbf{Buffering:} The implementation of the $3 \times 3$ convolution introduces an input data access pattern which requires multiple lines of the input. Because $FM_i$ data is streamed in the accelerator, $FM_i$ data lines must be buffered. In the general case, the size of the input tile buffer $ITB$ is:
\begin{equation}
\label{eq:bufin}
ITB = K \cdot X_i \cdot C_i,
\end{equation}
where $K$ denotes the kernel size (considering that $K=K_h=K_w$, see Figure \ref{fig:conv3d}), and $X_i$ and $C_i$ denote the width and the channels of $FM_i$ respectively. In the SqJ case, where support for up to $3 \times 3 \times C_i$ 3D kernels is required, $K=3$ and $ITB_{3 \times 3}$ is implemented as a set of 3 line buffers whose access is determined by a pointer array. In this way, $ITB_{3 \times 3}$ shifts down the $FM_i$ without the need for any data shift to take place; only the lowest, as defined by the pointer array, line buffer gets updated. This shift mechanism is also used by the input tile window buffer $ITWB$ (depicted as $IW$ in Figure \ref{fig:conv3d}) to update only one of its columns as it shifts horizontally on the $ITB$, taking advantage of the spatial locality of the input data. Figure \ref{fig:itb} shows the internal organization of $ITB$ and the operation of pointer array for $ITB_{3 \times 3}$. Apart from the $ITB$ and $ITWB$, buffers are used to store the weights, the bias, and one pixel of $FM_o$. The buffer used to store the $FM_o$ pixel could be omitted if each output channel was calculated serially, but buffering is required to calculate multiple output channels in parallel and to stream them out of the accelerator in order.
\\
\textbf{Parallelism exploitation:} SqJ takes advantage of the fact that SqN increases in the input channel dimension and, with the exception of the first conv layer, all conv layers' input channels have a GCD equal to $CI_{min}=16$. The accelerator is designed to perform $CI_{min}$ multiplications concurrently. These $CI_{min}$ products are then fed to an accumulator unit which outputs a $CI_{min}$ MAC result. The combination of the $CI_{min}$ concurrent multiplications plus the accumulator unit forms a MAC-$CI_{min}$ unit which is pipelined. $CI_{min}$ is a design parameter and can be easily modified according to the architecture of a different DCNN. This intra-kernel parallelism has the advantage that it exploits parallelism in the input channel dimension and it is independent from the kernel size $K$. Thus, SqJ can be easily modified to support kernel sizes larger than $3 \times 3$. Another form of parallelism that is used is the concurrent calculation of multiple output channels for a specific output pixel. This is achieved by splitting the weights buffer in $2^n$ ($n=1, \, 2, \, 3, \, ...$) equal groups of 3D kernels and assigning them to $2^n$ MAC-$CI_{min}$ units.
\\
\\
\textbf{Operation:} First step in the operation of SqJ is the initialization of the input buffers. Weights and bias are brought from off-chip memory and, only in the case where kernel $K=3$, the $ITB$ is initialized. After the initialization step, the convolution begins:
\begin{itemize}
\item[$\bullet$] For each row of $FM_o$: (1) only if $K=3$, the $ITB$ is shifted down (in $FM_i$) and two $FM_i$ pixels are written in the empty line buffer, and (2) only if $K=3$, the $ITWB$ is initialized with $ITB$ data.
\item[$\bullet$] For each column of each row of $FM_o$: (1) $ITB$ is updated with a new $FM_i$ pixel and $ITWB$ is updated with a new $ITB$ column, (2) the weight buffers and the $ITWB$ are used to calculate one pixel of $FM_o$, and (3) the computed pixel is written back to off-chip memory.
\end{itemize}

\subsection{Implementation}
\label{sec:sqj_impl}

\begin{figure}[]
    \centering
    \fbox{ \includegraphics[scale=0.25]{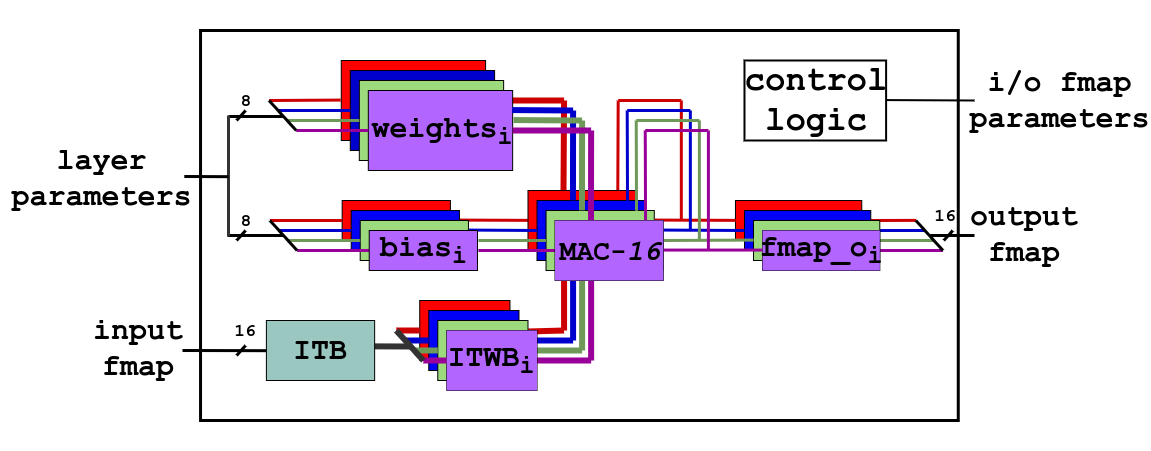} }
    \caption[]{SqJ block diagram implemented with 4 MAC-$CI_{min}$ units. Bold lines denote $CI_{min}=16$ times the data size shown at the left side of the figure.}
    \label{fig:sqj}
\end{figure}

FPGA algorithm acceleration is not as trivial as implementing an algorithm in SW using a general purpose programming language such as C/C++. Even though HLS tools advertise the automatic generation of FPGA IP cores from C/C++ code, this process requires knowledge of the architecture of the FPGA device, knowledge of the internals of the HLS compiler \cite{hls}, and use of a C/C++ coding style compatible with the HLS capabilities. This paragraph describes the process of generating an IP core for SqJ using the Xilinx VHLS tool and implementing it as a real application using the SDSoC tool.
\\
\\
\textbf{Coding style:} Hardware description languages (HDL) books warn the reader that if the designer cannot understand what logic circuit is described by the HDL code, then the design tool is not likely to synthesize the circuit that the designer is trying to model \cite{hdl}. The same applies for the C/C++ code used as input to VHLS. A result of this coding style is the implementation of $ITB$ shown in Figure \ref{fig:itb}, which uses the HW model of pointers to the rows of a two-dimensional array. Even though VHLS simplifies the HW design of an algorithm, it doesn't provide a straightforward way for making a design scalable as it is the case with the combination of generate constructs and generics/parameters used in HDLs.
\\
\\
\textbf{Interfaces:} The SqJ IP core requires buffers for the weights, the bias, the $ITB$ ($FM_i$), and the $FM_o$ buffer for storing the output pixel. Three FIFO interfaces are used to stream data in and out of the IP core; one for streaming in the parameter (weights, bias) data, one for the $FM_i$ data, and one for the output ($FM_o$) data. In addition, an AXI-Lite interface is used for acquiring the rest of the required HW function arguments. The SDSoC tool is used for interface synthesis. 
\\
\\
\textbf{Optimizations:} VHLS provides many optimization possibilities both in terms of performance and resource usage \cite{arc-hls}. 
\begin{itemize}
\item[$\bullet$] \textbf{Parallelism:} SqJ exploits parallelism in: (a) the input channel dimension (intra-kernel parallelism), by calculating the result of $CI_{min}$ MACs every clock cycle of the operation of the pipelined MAC-$CI_{min}$ unit, and (b) the output channel dimension, by calculating $2^n$ ($n=1, \, 2, \, 3, \, ...$) output channels concurrently. Parallelism in (a) requires a $CI_{min}$-wide data register and partitioning the operand buffers (array partitioning) in a way which makes them able to provide $CI_{min}$ outputs concurrently. Parallelism in (b) requires $2^n$ $ITWB$ buffers and the same number of MAC-$CI$ units.
\item[$\bullet$] \textbf{Arbitrary precision types:} To further decrease the model size of SqN and reduce the amount of logic required by SqJ, fixed-point quantization in both the parameters and the $FM_i$ is used. Specifically, Ristretto \cite{ris} is used to specify the proper quantization of the parameters (weights and bias) and the $FM_i$. Parameters are quantized at 8 bits (1 bit integer + 7 bits fractional) and $FM_i$ at 16 bits (13 bits integer + 3 bits fractional), achieving 0.88\% top-5 accuracy loss without performing any fine-tuning. 
\end{itemize}
In Figure \ref{fig:sqj}, the block diagram of SqJ, implemented (for simplicity) with 4 MAC-$CI_{min}$ units, is shown. Since the parallelization factor is equal to 4, the sizes of the buffers are: (1.179648/4) Mbits for the \texttt{weights\textsubscript{i}}, (2048/4) bits for the \texttt{bias\textsubscript{i}}, 344.064 Kbits for the \texttt{ITB}, 73.728 Kbits for each \texttt{ITWB\textsubscript{i}}, and (4096/4) bits for the \texttt{fmap\_o\textsubscript{i}}. Table \ref{table1} presents the FPGA resources required for the implementation of \textbf{conv\_l0}, the accelerator of the first SqN conv layer, and \textbf{SqJ}, in an 8 MAC-16 unit configuration, on the XC7Z020 FPGA SoC. The \textbf{conv\_l0 + SqJ} implementation is the one used in the results of the next Section.

\begin{table}[]
\centering
\caption{Resource Utilization of conv\_l0 and SqJ on the XC7Z020 FPGA SoC}
\label{table1}

\scalebox{0.7}{
\begin{tabular}{|l|r|r|r|r|r|r|r|}
\hline
\rowcolor[HTML]{EFEFEF} 
\multicolumn{2}{|l|}{\cellcolor[HTML]{EFEFEF}\textbf{}}                                                   & \multicolumn{2}{c|}{\cellcolor[HTML]{EFEFEF}\textbf{conv\_l0}}                                                               & \multicolumn{2}{c|}{\cellcolor[HTML]{EFEFEF}\textbf{SqJ}}                                                                    & \multicolumn{2}{c|}{\cellcolor[HTML]{EFEFEF}\textbf{conv\_l0 + SqJ}}                                                         \\ \hline
\rowcolor[HTML]{EFEFEF} 
\textbf{Resource}                       & \multicolumn{1}{c|}{\cellcolor[HTML]{EFEFEF}\textbf{Available}} & \multicolumn{1}{c|}{\cellcolor[HTML]{EFEFEF}\textbf{Util.}} & \multicolumn{1}{c|}{\cellcolor[HTML]{EFEFEF}\textbf{Util. \%}} & \multicolumn{1}{c|}{\cellcolor[HTML]{EFEFEF}\textbf{Util.}} & \multicolumn{1}{c|}{\cellcolor[HTML]{EFEFEF}\textbf{Util. \%}} & \multicolumn{1}{c|}{\cellcolor[HTML]{EFEFEF}\textbf{Util.}} & \multicolumn{1}{c|}{\cellcolor[HTML]{EFEFEF}\textbf{Util. \%}} \\ \hline
\cellcolor[HTML]{EFEFEF}\textbf{LUT}    & \textbf{53200}                                                  & 9405                                                        & 17.678                                                         & 12692                                                       & 23.857                                                         & \textbf{20631}                                              & \textbf{38.780}                                                \\ \hline
\cellcolor[HTML]{EFEFEF}\textbf{LUTRAM} & \textbf{17400}                                                  & 707                                                         & 4.063                                                          & 726                                                         & 4.172                                                          & \textbf{1273}                                               & \textbf{7.316}                                                 \\ \hline
\cellcolor[HTML]{EFEFEF}\textbf{FF}     & \textbf{106400}                                                 & 15459                                                       & 14.529                                                         & 18114                                                       & 17.024                                                         & \textbf{30554}                                              & \textbf{28.716}                                                \\ \hline
\cellcolor[HTML]{EFEFEF}\textbf{BRAM}   & \textbf{140}                                                    & 13                                                          & 9.285                                                          & 124                                                         & 88.571                                                         & \textbf{134.5}                                              & \textbf{96.071}                                                \\ \hline
\cellcolor[HTML]{EFEFEF}\textbf{DSP}    & \textbf{220}                                                    & 37                                                          & 16.818                                                         & 149                                                         & 67.727                                                         & \textbf{186}                                                & \textbf{84.545}                                                \\ \hline
\end{tabular}
}
\end{table}

\section{Performance evaluation}
\label{sec:perf}

Table \ref{table2} presents the per-layer execution times, the accuracy, and the chip power consumption\footnote{\label{note1}In the case of the ARM Cortex-A53, we measure RPI3 board power consumption, because there is no way to acquire power consumption measurements or estimations for the Broadcom 2837 SoC.} of SqN implemented on 4 different processing unit configurations, an Intel Core i3-7100U@2.4GHz core (Intel NUC), an ARM Cortex-A53@1.2GHz core (Raspberry Pi 3 (RPI3) Model B V1.2), an ARM Cortex-A9@667MHz core (Xilinx ZC702), and an ARM Cortex-A9@667MHz core with the SqJ@100MHz accelerator in an 8 MAC-16 unit configuration (Xilinx ZC702).

\begin{table}[!h]
\centering
\caption{SqN Application Execution time / Accuracy / Power Results}
\label{table2}

\scalebox{0.7}{
\begin{tabular}{lrrrr}
\hline
\multicolumn{1}{|l|}{\cellcolor[HTML]{EFEFEF}\textbf{\begin{tabular}[c]{@{}c@{}}Processing Unit\end{tabular}}}                                                                                                   & \multicolumn{1}{c|}{\cellcolor[HTML]{EFEFEF}\textbf{\begin{tabular}[c]{@{}c@{}}NUC \\ Intel i3@2.4GHz\end{tabular}}} & \multicolumn{1}{c|}{\cellcolor[HTML]{EFEFEF}\textbf{\begin{tabular}[c]{@{}c@{}}RPI3 \\ ARM A53@1.2GHz\end{tabular}}} & \multicolumn{1}{c|}{\cellcolor[HTML]{EFEFEF}\textbf{\begin{tabular}[c]{@{}c@{}}ZC702 \\ ARM A9@667MHz\end{tabular}}} & \multicolumn{1}{c|}{\cellcolor[HTML]{EFEFEF}\textbf{\begin{tabular}[c]{@{}c@{}}ZC702\\ ARM A9@667MHz\\ \cellcolor{blue!25}conv\_l0@100MHz \\ SqJ@100MHz\end{tabular}}} \\ \hline

\multicolumn{5}{|c|}{\cellcolor[HTML]{AAAAAA}\textbf{SqN Implementation Accuracy (bits)}}                                                                                                                                                                                                                                                                                                                                                                                                                                                                                                                                              \\ \hline

\multicolumn{1}{|l|}{\cellcolor[HTML]{EFEFEF}\textbf{Activations}}                                                            
& \multicolumn{3}{c|}{32}                                                                                        
& \multicolumn{1}{c|}{16}                                                                                                   
\\ \hline

\multicolumn{1}{|l|}{\cellcolor[HTML]{EFEFEF}\textbf{Weights, Bias}}                                                            
& \multicolumn{3}{c|}{32}                                                                                        
& \multicolumn{1}{c|}{8}                                                                                                   
\\ \hline

\multicolumn{5}{|c|}{\cellcolor[HTML]{AAAAAA}\textbf{SqN Application Per-Layer Execution Time Results (ms)}}                                                                                                                                                                                                                                                                                                                                                                                                                                                                                                                                              \\ \hline
\multicolumn{1}{|l|}{\cellcolor[HTML]{EFEFEF}\textbf{Load Image}}                                                            
& \multicolumn{1}{r|}{0.1761}                                                                                        
& \multicolumn{1}{r|}{1.2137}                                                                                              
& \multicolumn{1}{r|}{21.3210}                                                                                    
& \multicolumn{1}{r|}{54.4263}                                                                                                   
\\ \hline
\multicolumn{1}{|l|}{\cellcolor[HTML]{EFEFEF}\textbf{0:Conv}}                                                            
& \multicolumn{1}{r|}{25.3118}                                                                                        
& \multicolumn{1}{r|}{131.5186}                                                                                              
& \multicolumn{1}{r|}{297.2426}                                                                                    
& \multicolumn{1}{r|}{\cellcolor{blue!25}26.2756}                                                                                                   
\\ \hline
\multicolumn{1}{|l|}{\cellcolor[HTML]{EFEFEF}\textbf{1:Maxpool}}                                                         
& \multicolumn{1}{r|}{2.0531}                                                                                         
& \multicolumn{1}{r|}{18.2868}                                                                                              
& \multicolumn{1}{r|}{28.7206}                                                                                     
& \multicolumn{1}{r|}{22.7574}                                                                                                   
\\ \hline
\multicolumn{1}{|l|}{\cellcolor[HTML]{EFEFEF}\textbf{2:Fire}}                                                            
& \multicolumn{1}{r|}{16.1473 }                                                                                        
& \multicolumn{1}{r|}{142.7623 }                                                                                              
& \multicolumn{1}{r|}{446.1214 }                                                                                    
& \multicolumn{1}{r|}{32.6526}                                                                                                   
\\ \hline
\multicolumn{1}{|l|}{\cellcolor[HTML]{EFEFEF}\textbf{3:Fire}}                                                            
& \multicolumn{1}{r|}{17.0744 }                                                                                        
& \multicolumn{1}{r|}{150.7194 }                                                                                              
& \multicolumn{1}{r|}{474.1013 }                                                                                    
& \multicolumn{1}{r|}{34.7981}                                                                                                   
\\ \hline
\multicolumn{1}{|l|}{\cellcolor[HTML]{EFEFEF}\textbf{4:Maxpool}}                                                         
& \multicolumn{1}{r|}{1.3333}                                                                                         
& \multicolumn{1}{r|}{13.4446}                                                                                               
& \multicolumn{1}{r|}{27.3646}                                                                                     
& \multicolumn{1}{r|}{18.0916}                                                                                                  
 \\ \hline
\multicolumn{1}{|l|}{\cellcolor[HTML]{EFEFEF}\textbf{5:Fire}}                                                            
& \multicolumn{1}{r|}{13.5606 }                                                                                         
& \multicolumn{1}{r|}{124.2315 }                                                                                              
& \multicolumn{1}{r|}{450.0168 }                                                                                    
& \multicolumn{1}{r|}{17.7738}                                                                                                   
\\ \hline
\multicolumn{1}{|l|}{\cellcolor[HTML]{EFEFEF}\textbf{6:Fire}}                                                            
& \multicolumn{1}{r|}{14.5805 }                                                                                        
& \multicolumn{1}{r|}{135.3108 }                                                                                              
& \multicolumn{1}{r|}{482.2875 }                                                                                    
& \multicolumn{1}{r|}{18.9882}                                                                                                   
\\ \hline
\multicolumn{1}{|l|}{\cellcolor[HTML]{EFEFEF}\textbf{7:Maxpool}}                                                         
& \multicolumn{1}{r|}{0.6023}                                                                                        
& \multicolumn{1}{r|}{7.1370}                                                                                               
& \multicolumn{1}{r|}{14.4114}                                                                                     
& \multicolumn{1}{r|}{9.4158}                                                                                                    
\\ \hline
\multicolumn{1}{|l|}{\cellcolor[HTML]{EFEFEF}\textbf{8:Fire}}                                                            
& \multicolumn{1}{r|}{7.4712 }                                                                                         
& \multicolumn{1}{r|}{69.1218 }                                                                                              
& \multicolumn{1}{r|}{257.9832 }                                                                                    
& \multicolumn{1}{r|}{8.6426}                                                                                                    
\\ \hline
\multicolumn{1}{|l|}{\cellcolor[HTML]{EFEFEF}\textbf{9:Fire}}                                                           
& \multicolumn{1}{r|}{7.8755 }                                                                                         
& \multicolumn{1}{r|}{72.4013 }                                                                                              
& \multicolumn{1}{r|}{273.4599 }                                                                                    
& \multicolumn{1}{r|}{8.8704}                                                                                                    
\\ \hline
\multicolumn{1}{|l|}{\cellcolor[HTML]{EFEFEF}\textbf{10:Fire}}                                                           
& \multicolumn{1}{r|}{13.1197 }                                                                                        
& \multicolumn{1}{r|}{125.8514 }                                                                                              
& \multicolumn{1}{r|}{497.6390 }                                                                                    
& \multicolumn{1}{r|}{12.2322}                                                                                                   
\\ \hline
\multicolumn{1}{|l|}{\cellcolor[HTML]{EFEFEF}\textbf{11:Fire}}                                                           
& \multicolumn{1}{r|}{13.6331 }                                                                                        
& \multicolumn{1}{r|}{132.5349 }                                                                                              
& \multicolumn{1}{r|}{517.09514 }                                                                                    
& \multicolumn{1}{r|}{12.7946}                                                                                                   
\\ \hline
\multicolumn{1}{|l|}{\cellcolor[HTML]{EFEFEF}\textbf{12:Conv}}                                                           
& \multicolumn{1}{r|}{34.7681 }                                                                                        
& \multicolumn{1}{r|}{324.9181 }                                                                                              
& \multicolumn{1}{r|}{1257.4682 }                                                                                   
& \multicolumn{1}{r|}{33.9618}                                                                                                   
\\ \hline
\multicolumn{1}{|l|}{\cellcolor[HTML]{EFEFEF}\textbf{13:Fixed2float}}                                                           
& \multicolumn{1}{r|}{0.0001}                                                                                        
& \multicolumn{1}{r|}{0.0004}                                                                                              
& \multicolumn{1}{r|}{0.0003}                                                                                   
& \multicolumn{1}{r|}{15.4479}                                                                                                   
\\ \hline
\multicolumn{1}{|l|}{\cellcolor[HTML]{EFEFEF}\textbf{13:Avgpool}}                                                        
& \multicolumn{1}{r|}{1.5295}                                                                                         
& \multicolumn{1}{r|}{4.3149}                                                                                               
& \multicolumn{1}{r|}{5.7796}                                                                                      
& \multicolumn{1}{r|}{5.7085}                                                                                                    
\\ \hline
\multicolumn{1}{|l|}{\cellcolor[HTML]{EFEFEF}\textbf{14:Softmax}}                                                        
& \multicolumn{1}{r|}{0.0260}                                                                                         
& \multicolumn{1}{r|}{0.1528}                                                                                               
& \multicolumn{1}{r|}{0.2212}                                                                                      
& \multicolumn{1}{r|}{0.2220}                                                                                                    
\\ \hline
\rowcolor[HTML]{FFFFFF} 
\multicolumn{1}{|l|}{\cellcolor[HTML]{C0C0C0}\textbf{Total Conv}}                                                   
& \multicolumn{1}{r|}{\cellcolor[HTML]{FFFFFF}\textbf{162.4322}}                                                      
& \multicolumn{1}{r|}{\cellcolor[HTML]{FFFFFF}\textbf{1395.4275}}                                                            
& \multicolumn{1}{r|}{\cellcolor[HTML]{FFFFFF}\textbf{4892.9386}}                                                  
& \multicolumn{1}{r|}{\cellcolor[HTML]{FFFFFF}\textbf{174.9867}}                                                                 
\\ \hline
\rowcolor[HTML]{FFFFFF} 
\multicolumn{1}{|l|}{\cellcolor[HTML]{C0C0C0}\textbf{Total Merge}}                                                   
& \multicolumn{1}{r|}{\cellcolor[HTML]{FFFFFF}\textbf{13.74}}                                                      
& \multicolumn{1}{r|}{\cellcolor[HTML]{FFFFFF}\textbf{13.89}}                                                            
& \multicolumn{1}{r|}{\cellcolor[HTML]{FFFFFF}\textbf{60.43}}                                                  
& \multicolumn{1}{r|}{\cellcolor[HTML]{FFFFFF}\textbf{31.97}}                                                                 
\\ \hline
\rowcolor[HTML]{FFFFFF} 
\multicolumn{1}{|l|}{\cellcolor[HTML]{C0C0C0}\textbf{Total Maxpool}}                                                   
& \multicolumn{1}{r|}{\cellcolor[HTML]{FFFFFF}\textbf{3.99}}                                                      
& \multicolumn{1}{r|}{\cellcolor[HTML]{FFFFFF}\textbf{38.87}}                                                            
& \multicolumn{1}{r|}{\cellcolor[HTML]{FFFFFF}\textbf{70.49}}                                                  
& \multicolumn{1}{r|}{\cellcolor[HTML]{FFFFFF}\textbf{50.26}}                                                                 
\\ \hline
\rowcolor[HTML]{FFFFFF} 
\multicolumn{1}{|l|}{\cellcolor[HTML]{C0C0C0}\textbf{Total}}                                                             
& \multicolumn{1}{r|}{\cellcolor[HTML]{FFFFFF}\textbf{169.2627}}                                                     
& \multicolumn{1}{r|}{\cellcolor[HTML]{FFFFFF}\textbf{1453.9202}}                                                            
& \multicolumn{1}{r|}{\cellcolor[HTML]{FFFFFF}\textbf{5051.2337}}                                                  
& \multicolumn{1}{r|}{\cellcolor[HTML]{FFFFFF}\textbf{333.0595}}                                                                 
\\ \hline

\rowcolor[HTML]{FFFFFF} 
\multicolumn{1}{|l|}{\cellcolor[HTML]{C0C0C0}\textbf{FPS}}                                                             
& \multicolumn{1}{r|}{\cellcolor[HTML]{FFFFFF}\textbf{5.907}}                                                     
& \multicolumn{1}{r|}{\cellcolor[HTML]{FFFFFF}\textbf{0.687}}                                                            
& \multicolumn{1}{r|}{\cellcolor[HTML]{FFFFFF}\textbf{0.198}}                                                  
& \multicolumn{1}{r|}{\cellcolor[HTML]{FFFFFF}\textbf{3.002}}                                                                 
\\ \hline

\multicolumn{5}{|c|}{\cellcolor[HTML]{AAAAAA}\textbf{SqN ILSVRC12 Accuracy Results (\%)}}                                                                                                                                                                                                                                                                                                                                                                                                                                                                                                                                                      \\ \hline

\multicolumn{1}{|l|}{\cellcolor[HTML]{EFEFEF}\textbf{Top-1}}       
& \multicolumn{3}{c|}{58.38}                                                                                           
& \multicolumn{1}{c|}{57.46}                                          
\\ \hline

\multicolumn{1}{|l|}{\cellcolor[HTML]{EFEFEF}\textbf{Top-5}}       
& \multicolumn{3}{c|}{81.01}                                                                                           
& \multicolumn{1}{c|}{\textbf{80.29}}                                          
\\ \hline

\multicolumn{5}{|c|}{\cellcolor[HTML]{AAAAAA}\textbf{SqN Application CPU/SoC Power Consumption Results (Watts)}}                                                                                                                                                                                                                                                                                                                                                                                                                                                                                                                                                      \\ \hline

\multicolumn{1}{|l|}{\cellcolor[HTML]{EFEFEF}\textbf{Technology}}       
& \multicolumn{1}{r|}{14nm}                                                                                           
& \multicolumn{1}{r|}{n/a}                                                                                                 
& \multicolumn{1}{r|}{28nm}                                                                                        
& \multicolumn{1}{r|}{28nm}                                          
\\ \hline

\multicolumn{1}{|l|}{\cellcolor[HTML]{EFEFEF}\textbf{Chip Power}}       
& \multicolumn{1}{r|}{5.3253}                                                                                           
& \multicolumn{1}{r|}{2.9}                                                                                                 
& \multicolumn{1}{r|}{1.569}                                                                                        
& \multicolumn{1}{r|}{\textbf{2.275}}                                          
\\ \hline

\rowcolor[HTML]{FFFFFF} 
\multicolumn{1}{|l|}{\cellcolor[HTML]{C0C0C0}\textbf{FPS/W}}                                                             
& \multicolumn{1}{r|}{\cellcolor[HTML]{FFFFFF}\textbf{1.109}}                                                     
& \multicolumn{1}{r|}{\cellcolor[HTML]{FFFFFF}\textbf{0.237}}                                                            
& \multicolumn{1}{r|}{\cellcolor[HTML]{FFFFFF}\textbf{0.126}}                                                  
& \multicolumn{1}{r|}{\cellcolor[HTML]{FFFFFF}\textbf{1.319}}                                                                 
\\ \hline

\end{tabular}
}
\end{table}
\par
SqN is a single floating point precision C/C++ Linux application accelerated with single-instruction multiple-data (SIMD) instruction set extensions (Intel AVX, ARM NEON) and executed on a single core of the target CPU-only processing systems. In the case where the SqJ accelerator is used, the implementation uses 16 bits for the activations and 8 bits for the weights and bias. GCC (version 6.3.0 for the Intel (64-bit) and RPI3 (32-bit) configurations, and version 6.2.1 for the ZC702 (32-bit) configuration) with the -O3 flag is used to build the SqN Linux application. Execution times are an average of 1000 inference iterations. Power consumption is acquired: (1) using Intel PCM\footnote{\url{https://www.intel.com/software/pcm}} while the processing system executes 1000 SqN iterations, in the case of the Intel i3 CPU, (2) using a power plug and measuring board power consumption, in the case of RPI3, and (3) using Xilinx XPE\footnote{\url{https://www.xilinx.com/products/technology/power/xpe.html}} in the case of Xilinx ZC702. Accuracy is evaluated using the Ristretto\footnote{\url{https://github.com/pmgysel/caffe}} tool.
\par
Results show that the SqJ configuration achieves an 15.16x execution time speedup in SqN inference when compared to the ARM A9 core configuration, 4.36x execution time speedup in SqN inference when compared to the ARM A53 core, and similar convolution performance (see \textbf{Total Conv} in Table \ref{table2}) to the Intel i3 core configuration, with less than 1\% top-5 accuracy loss. In terms of performance per Watt, frames per second per Watt (FPS/W), the SqJ implementation is 10.46 times better than the ARM A9 core configuration; again, with less than 1\% accuracy loss. The \textbf{Load Image} execution time in the SqJ implementation includes the conversion of the image from 32-bit floating point to 16-bit fixed point; that's why it takes more than double of the ARM A9 corresponding time. Because of the use of lower precision for the activations, \textbf{Total Merge} (merge operations are included in the Fire layers) and \textbf{Total Maxpool} operations require much less time than the ARM A9 implementation. Furthermore, the Maxpool layers require 15\% of the \textbf{Total} SqJ implementation time and could be incorporated in a future SqJ implementation. Table \ref{table3} summarizes the characteristics of the SqJ implementation.

\begin{table}[!h]
\centering
\caption{SqJ (conv\_l0+SqJ) Implementation Summary}
\label{table3}
\begin{tabular}{|l|c|}
\hline
                                                                                                   & \multicolumn{1}{l|}{\cellcolor[HTML]{EFEFEF}\textbf{SqueezeNet v1.1}} \\ \hline
\cellcolor[HTML]{EFEFEF}\textbf{FPGA}                                                              & Zynq XC7Z020                                                          \\ \hline
\cellcolor[HTML]{EFEFEF}\textbf{Frequency (MHz)}                                                   & 100                                                                   \\ \hline
\cellcolor[HTML]{EFEFEF}\textbf{Design Tool}                                                       & Vivado HLS                                                            \\ \hline
\cellcolor[HTML]{EFEFEF}\textbf{DCNN Ops (GOPs)}                                                   & 0.7755                                                                \\ \hline
\cellcolor[HTML]{EFEFEF}\textbf{Precision}                                                         & 8-16 bits                                                             \\ \hline
\cellcolor[HTML]{EFEFEF}\textbf{DSP (Util.)}                                                       & 186 (84.5\%)                                                          \\ \hline
\cellcolor[HTML]{EFEFEF}\textbf{BRAM (Util.)}                                                      & 134.5 (96\%)                                                          \\ \hline
\cellcolor[HTML]{EFEFEF}\textbf{LUT (Util.)}                                                       & 20631 (38.8\%)                                                        \\ \hline
\cellcolor[HTML]{EFEFEF}\textbf{LUTRAM (Util.)}                                                    & 1273 (7.3\%)                                                          \\ \hline
\cellcolor[HTML]{EFEFEF}\textbf{FF (Util.)}                                                        & 30554 (28.7\%)                                                       
\\ \hline
\cellcolor[HTML]{EFEFEF}\textbf{Conv Latency/Image (ms)}                                           & 175                                                                   \\ \hline
\cellcolor[HTML]{EFEFEF}\textbf{Throughput (GOPs)}													 & 4.43                                                                  \\ \hline
\cellcolor[HTML]{EFEFEF}\textbf{\begin{tabular}[c]{@{}l@{}}Top-5\\ ILSVRC12 Accuracy\end{tabular}} & 80.29\%                                                               \\ \hline
\end{tabular}
\end{table}

\section{Conclusion}
\label{sec:con}

In this paper, we present the design and the implementation of SqJ, an FPGA-based convolution layer accelerator which can be used to boost the performance of an embedded mobile processor running a CV task. The accelerator, consisting of a buffering architecture and multiple computational units, is designed using the Xilinx Vivado HLS tool. The Ristretto tool is used to squeeze the SqN DCNN in the Xilinx XC7Z020 FPGA SoC, and the Xilinx SDSoC tool is used to deploy SqJ accelerated SqN to the XC7Z020 device. To the best of our knowledge, our work is the first one which implements the SqN DCNN in a small FPGA SoC device, such as the XC7Z020, and achieves 80.29\% top-5 ILSVRC12 accuracy (using XC7Z020). Results show that SqJ accelerates by 15.16 times the SqN inference execution time of an embedded mobile processor while being 10.46 times more power efficient with less than 1\% top-5 accuracy drop. Improvements to the HLS SqJ design could include: (1) Maxpool layer support, since they require considerable amount (15\%) of the total inference time on a mobile ARM core, and (2) streaming execution, to avoid memory accesses for fmaps (requires additional BRAM resources). Future work could use an enhanced version of SqJ as a template and perform multiobjective optimization for finding the best solution in terms of performance, resources, accuracy, power, and cost.

%

%
\end{document}